\documentclass[runningheads]{llncs}

\usepackage{eccv}

\usepackage{eccvabbrv}

\usepackage{graphicx}
\usepackage{booktabs}

\usepackage[accsupp]{axessibility}  % Improves PDF readability for those with disabilities.

\usepackage{hyperref}

\usepackage{orcidlink}

\usepackage[dvipsnames]{xcolor}

\newcommand{\para}[1]{\smallskip \noindent \textbf{{#1}.}}

\newcommand{\pred}{\mathcal{P}} % predictor
\newcommand{\data}{\mathcal{D}} % dataset
\newcommand{\predhor}{T} % prediction horizon
\newcommand{\histhor}{H} % history horizon
\newcommand{\ptime}{\tau} % predictor training time
\newcommand{\nagents}{N}
\newcommand{\adapt}{\mathcal{A}} %adaptation rule

\newcommand{\sid}{k} % scene id
\newcommand{\snum}{K} % number of scenes

\newcommand{\heatmap}{\mathbf{M}}

\usepackage{newunicodechar}
\newunicodechar{ }{\,}  % Replace the narrow no-break space with a small space.
\begin{document}

\title{Adaptive Human Trajectory Prediction \\ via Latent Corridors} 


% Second Author\inst{2,3}\orcidlink{1111-2222-3333-4444} \and
% Third Author\inst{3}\orcidlink{2222--3333-4444-5555}}

\author{Neerja Thakkar\inst{1} \and Karttikeya Mangalam\inst{1} \and Andrea Bajcsy\inst{2} \and Jitendra Malik\inst{1}}

\authorrunning{N. Thakkar et al.}

% Springer Heidelberg, Tiergartenstr.~17, 69121 Heidelberg, Germany

\institute{UC Berkeley \and Carnegie Mellon University}

\maketitle

\begin{abstract}
% \vspace{-0.99em}
Human trajectory prediction is typically 
posed as a zero-shot generalization problem: a predictor is learnt on a dataset of human motion in training scenes, and then deployed on unseen test scenes. 
While this paradigm has yielded tremendous progress, it fundamentally assumes that trends in human behavior within the deployment scene are constant over time. 
As such, current prediction models are unable to adapt to transient human behaviors, such as crowds temporarily gathering to see buskers, pedestrians hurrying through the rain and avoiding puddles, or a protest breaking out. 
We formalize the problem of context-specific adaptive trajectory prediction and propose a new adaptation approach inspired by prompt tuning called latent corridors. 
By augmenting the input of a pre-trained human trajectory predictor with learnable image prompts, the predictor improves in the deployment scene by inferring trends from extremely small amounts of new data (e.g., 2 humans observed for 30 seconds).  
With less than $0.1\%$ additional model parameters, we see up to $23.9\%$ ADE improvement in MOTSynth simulated data and $16.4\%$ ADE in MOT and Wildtrack real pedestrian data.
Qualitatively, we observe that latent corridors imbue predictors with an awareness of scene geometry and context-specific human behaviors that non-adaptive predictors struggle to capture. 
% \vspace{-0.99em}

\end{abstract}
\section{Introduction}
\label{sec:intro}

\begin{figure*}[t]
  \centering
    \includegraphics[width=0.99\textwidth,clip]{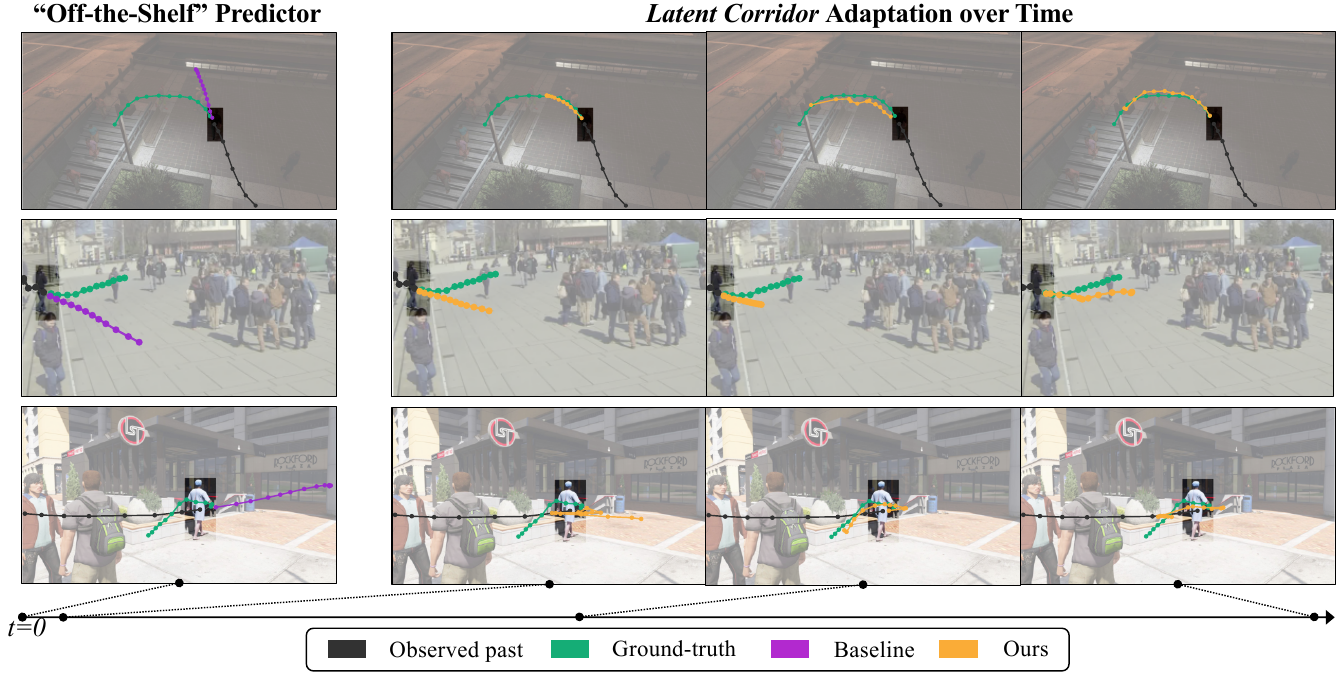}
    
    \vspace{-0.05in}
   \captionof{figure}{\textbf{Adaptive trajectory prediction.} (left) Given a history of human behavior (shown in black), the pre-trained predictor $\pred$ is unable to understand deployment scene-specific behavior trends, like people entering a subterranean subway entrance (bottom row) or mostly choosing to traverse the staircase as opposed to exploring other parts of the scene at nighttime (top row). (right) 
   When adapting, the number of people and amount of time determine the total number of trajectories observed, and we denote this time-dependent quantity human-seconds. Here, the three columns correspond to our method trained for a very small (left), medium (middle), and large amount of human-seconds (right).
    Our adaptive \textit{latent corridors} approach enables $\pred$ to quickly learn context-specific trends, improving predictions with even small amounts of data, and closing the gap between the ground-truth (green) and predicted behavior (orange).
   For example, in the middle row, $\pred$ predicts the person will move towards the camera, but as our method sees more human-seconds of data, it adapts to the trend that at this point of scene capture in the plaza, people tend to avoid the center of the plaza and instead move diagonally across it. }  
   % \vspace{-0.22in}
\label{fig:teaser}
  \vspace{-1.5em}
\end{figure*}

Human motion prediction is a fundamental skill for intelligent systems to effectively navigate the world, assist end-users, and visually survey a scene. To date, learning-based human trajectory prediction has been extensively studied, making huge strides in predicting multimodal future behavior \cite{mangalam2020not}, modeling multi-agent social interactions \cite{alahi2016social, ivanovic2018generative, salzmann2020trajectron++}, and accounting for scene context \cite{zhao2019multi, mangalam2021goals}. Some approaches have exploited the fact that humans tend to move in similar patterns by storing prior trajectories in a bank and matching a current observed trajectory to the closest stored one~\cite{meng2022forecasting,marchetti2020mantra,xu2022remember,yang2022continual}.

However, a fundamental assumption underlies the current trajectory prediction paradigm: predictors are assumed to be deployed within an \textit{unchanging} context. Nevertheless, in the real world, the context---and thus patterns of human behavior---will inevitably change over time, as Whyte observed in his famous study of urban public spaces~\cite{Whyte_1980}. For example, a new subway entrance being built causes people to descend and exit in new patterns (bottom row, Fig.~\ref{fig:teaser}), and icy sidewalks in the winter will be traversed differently than clear ones in the summer. These changes happen on shorter temporal horizons, too: Humans rushing to work during the day will largely ignore each other while nighttime party-goers will walk in cliques. A college campus will be quiet while class is in session and chaotic in the 10 minute period where students rush between classes. 
% These changes happen in many forms: a new subway entrance being built causes people to descend and exit in new patterns (bottom row, Fig.~\ref{fig:teaser}), icy sidewalks change how people move, and those rushing to work during the day largely ignore each other while nighttime party-goers walk in cliques. 
% it starts to rain or snow, or there is a musician who starts to play, or there is a protest. 
Even if data is captured from exactly the same location, human behavior patterns can change in minutes or hours. When faced with such changes, the performance of existing trajectory predictors immediately degrades.

Here, we study how to efficiently \textit{adapt} pre-trained trajectory predictors to observed human behavior within a deployment scene (right, Fig.~\ref{fig:teaser}). 
% We focus on adaptation that can be done in a data-efficient way and can preserve any useful structure within the base predictor in the new context (e.g., even if a new subway entrance lets people descend into the ground, other people may still move towards the existing above-ground coffee shops). 
Our technical approach takes inspiration 
from the recently popularized paradigm of \textit{prompt-tuning} in large language models \cite{lester2021power}, but instantiates this idea within the adaptive trajectory prediction problem. 
Specifically, we augment the input to an existing human trajectory predictor with a set of latent image prompts, one per each instance of a deployment scene we wish to adapt to. 
We leave all or most of the predictor frozen, and tune each input prompt with the same trajectory prediction loss that is used to train the base predictor, letting cues in human behaviour be captured while preserving useful structure in the original predictor (e.g., even if a new subway entrance lets people descend into the ground, other people may still move towards the existing above-ground coffee shops).
Our adaptation is data-efficient: even extremely small amounts of new human trajectories (e.g., $< 1$ minute of human seconds, corresponding to 2 humans observed over 30 seconds) can leak sufficient information about how humans interact with the scene (e.g., new subway entrance) or with each other (e.g., nighttime social cliques) to learn this prompt and improve future trajectory prediction.  
We call our learnable prompt a \textit{latent corridor}: a non-physical corridor that guides human behavior in this scene.

% Our experiments involve training latent corridors on scene snapshots over time. 
Given the small amounts of data, latent corridors train quickly---on average within minutes on a single 2080 Ti GPU---even without optimizations to improve training speed.
With more compute resources and training optimization, latent corridors hold the potential to learn in real time, adapting to transient events such a fire starting or a musician busking. 
Even with minimal compute and optimization, we also find useful \textit{offline} applications of latent corridors for adaptation to repeated events. For example, imagine analyzing pedestrian flow in a downtown business district --- it will differ during the rush hour before work on the weekday and on a weekend morning. 
% Additionally, every camera will capture a distinct 3D environment. 
During the course of a week, latent corridors could be trained to adapt to the varying scene and context specific behaviours present on different days, times, and surveillance camera viewpoints. Once training is complete, adapted models could be used \textit{online} for months, until an event such as construction starting or a change of season causes a significant change in behaviour. 
Our method is lightweight, making swapping latent corridors for different times of day or locations inexpensive, even on device.

In our experiments, learning latent corridors for prediction adaptation enables $23.9\%$ prediction accuracy improvement on simulated data from MOTSynth \cite{fabbri2021motsynth} compared to a non-adaptive predictor that takes as input the scene and has seen all the same training data. On real data from MOT and WildTrack \cite{chavdarova2018wildtrack}, we see a $16.4\%$ improvement on non-adaptive predictors and an $11.2\%$ improvement with adapting via our method as opposed to just fine-tuning, and on in-the-wild webcam data we see respective improvements of $26.8\%$ and $20.0\%$.
Latent corridor adaptation can improve prediction performance even with a very small amount of data, and continually improves as more data is observed over time. 
% Latent corridor adaptation can improve prediction performance even with a very small amount of data (e.g. a couple of humans for a few seconds), and continually improves as more data (e.g., many humans for a few minutes) is observed over time. 
% Qualitatively, we observe that latent corridors can inject 3D ground plane awareness into prediction performed from an arbitrary camera view angle, can help the predictor adapt to obstacles or occlusions in the scene, and learn context-specific patterns of human behavior. 

To summarize, our contributions are as follows:
\begin{enumerate}
    \item We formalize the adaptive trajectory prediction problem.
    \item We propose a novel method for adaptive trajectory prediction by learning lightweight latent corridor prompts in image space, outperforming non-adaptive trajectory predictors with less than a $0.1\%$ parameter increase.
    \item We demonstrate qualitative and quantitative improvements of up to $23.9\%$ ADE improvement in MOTSynth simulated data, $16.4\%$ ADE in MOT and Wildtrack real pedestrian data, and $26.8\%$ ADE on in-the-wild webcam data, over a scene-aware baseline. 
\end{enumerate}
\section{Related Work}
\label{sec:related_work}

%-------------------------------------------------------------------------
% \subsection{Human Trajectory Prediction}

\textbf{Human Trajectory Prediction} is a well-studied problem with a long and rich history~\cite{rudenko2020human}. Prediction methods started from simple models such as social forces~\cite{599328, helbing1995social}, and later added multimodality through Gaussian processes~\cite{tay2008modelling}. Recently, modern learning-based predictors have made significant progress in incorporating social interactions between humans modelled via RNNs~\cite{vemula2018social, ivanovic2019trajectron, salzmann2020trajectron++}, GANs~\cite{gupta2018social}, or conditional VAEs~\cite{mangalam2020not}, all while generating multimodal future trajectories. 
% to generate multiple feasible future trajectory predictions given an observed history and modeling of social dynamics. 
Another line of work incorporates the scene context in predictions, getting neural network features from a scene map~\cite{kitani2012activity, varshneya2017human, 8354239, lee2017desire, sadeghian2019sophie,marchetti2020mantra, manh2018scene}, embedding gridded scene context in an LSTM~\cite{manh2018scene,xue2018ss}, and more recently using transformers~\cite{ngiam2021scene,salzmann2023hst}. 
A few works incorporate scene context by projecting the trajectories into heatmaps so that the network can reason in image space~\cite{mangalam2021goals,cao2020long}, allowing for longer-term predictions. Our work builds upon these advances by adapting an RGB scene-aware pre-trained trajectory predictor \cite{mangalam2021goals} over time.

%-------------------------------------------------------------------------
% \subsection{Adapting Human Trajectory Predictors}
\para{Adapting Human Trajectory Predictors}
% As human trajectory predictors become more widespread, 
In the past few years, there has been a growing interest in lightweight ways to modify pre-trained predictors to new data. 
% in addition to improving predictors, works in recent years have also explored the idea of somehow adapting them. 
The key differences between these works lie in 1) \textit{what} they are adapting to, and 2) \textit{how} they adapt. 
Several works focus on cross-domain transfer, wherein a predictor trained for trajectory prediction in domain A (e.g., New York) is adapted to work in domain B (e.g., London). These works leverage architectures that partition generic trajectory prediction from domain-specific features~\cite{xu2022adaptive, wang2022transferable}, leverage adaptive meta-learning via Kalman flitering~\cite{ivanovic2023expanding}, or simply finetune the prediction network on new data~\cite{li2023pre}. 
Other works focus on online adaptation over time, for example by adapting to different agent dynamics using recursive least squares~\cite{abuduweili2019adaptable, cheng2019human}.
A few approaches carry out continual learning by storing embedded past/future trajectory pairs in a memory bank~\cite{marchetti2020mantra, xu2022remember, yang2022continual} or clustering them~\cite{sun2021three}, predicting by matching an individuals history to the most similar stored past. 
Going a step further,~\cite{meng2022forecasting} instead stores representative group trajectories in a scene and during inference, refines the selected trajectory with scene segmentation. In contrast, our novel prompting-based adaptation approach is memory-efficient, using constant extra parameter space regardless of trajectory dataset size, and inherently utilizes the scene segmentation, not requiring extra refinement steps. It also directly augments an existing model as opposed to requiring a new, specialized architecture.

\para{Prompt Tuning in Language \& Vision}
% We use a prompt tuning approach as part of our adaptation rule. 
With the rise of large pre-trained models, efficient adaptation for downstream tasks has gained increasing interest.  
In the large language model domain, prompt tuning---wherein a few tunable tokens (i.e., prompt) are prepended to input text and tuned per downstream task---has been shown to be remarkably effective \cite{lester2021power, li2021prefix, liu2023pre}.  
In vision models, image prompts~\cite{wang2023review} have been introduced to adapt a vision model to new tasks~\cite{bahng2022exploring} or instruct an inpainting model to carry out various computer vision tasks~\cite{visprompt}. Recently, there have also been attempts at visual prompt tuning, showing that large vision transformers can benefit from utilizing tuned prompts for continual learning or transfer learning~\cite{jia2022visual,sandler2022fine, wang2022learning, nie2023pro}. 
While our task is different, we were inspired by the prompting scheme of~\cite{jia2022visual} that learned classification on a new dataset by introducing a trainable input space prompt and tuning the prompt along with the transformer predictor head. We use visual prompts as a lightweight way of adapting a trajectory predictor. To our knowledge, we are the first work to apply prompt tuning to human trajectory prediction.

\vspace{-.5em}

%-------------------------------------------------------------------------
% \subsection{Affordances}

% \begin{itemize}
%     \item Affordances from Human Videos as a Versatile Representation for Robotics~\cite{bahl2023affordances}
% \end{itemize}

\vspace{-.5em}
\section{Problem Formulation}
\label{sec:our_atp}

\vspace{-0.5 em}
Here, we present a formalization of our proposed adaptive trajectory prediction problem. 
We seek to generate accurate future trajectories of $\nagents$ agents in a scene. 
At any time $\ptime$, let $\mathbf{o}_\ptime := o_{\ptime-\histhor:\ptime} \in \mathcal{O}$ be the $\histhor$-step history of observations input into the predictor, 
$\mathbf{y}_{\ptime} := y_{\ptime+1:\ptime+\predhor} \in \mathcal{Y}$ be the ground-truth $\predhor$-step future trajectory, and $\hat{\mathbf{y}}_{\ptime} := \hat{y}_{\ptime+1:\ptime+\predhor} \in \mathcal{Y}$ be the prediction outputs. 
We assume access to a base predictor $\pred$ pre-trained on a human motion dataset consisting of past and ground-truth future trajectories, $\data = \{\mathbf{o}_i, \mathbf{y}_i\}^K_{i=1}$ to minimize a loss function on trajectories, $\ell$ (e.g., mean squared error).

\para{Adaptive Trajectory Prediction (ATP)} 
We aim to adapt $\pred$ to a deployment scene by observing new human trajectory data over a time interval. 
Let the dataset of \textit{new} human trajectory data be $\data'_{0:T}$ collected over the $T$ step time interval.
% The deployment scenes being adapted to in 
Here, $\data'$ can contain data from real or synthetic deployment scenes, a new physical environment or one from the pre-training dataset, a new or previously seen time period (e.g., nightime vs. daytime), and can be obtained by a new or previously seen camera pose. 
However, we assume that human pedestrians are present in the scene and they are navigating an outdoor environment. 
We adapt to scenes captured for a similar amount of time, between $T= 45$ seconds and $5$ minutes. 
We aim to adapt to scene-specific characteristics and fluctuating events beyond those that can be exploited by conditioning only on a segmentation map. 

In the adaptive trajectory prediction problem, a subset of the deployment dataset $\data'_{0:t}$, $t < T$ is observed. 
The goal is create an adapted model $\adapt[\pred]$ such that the prediction error decreases over the future observed deployment data $\data'_{t:T}$: 
\begin{equation}
\ell\big(\adapt[\pred(\mathbf{o})], \mathbf{y}\big) < \ell\big(\pred(\mathbf{o}), \mathbf{y}\big), \quad (\mathbf{o}, \mathbf{y}) \in \data'_{t:T}.
\end{equation}
Intuitively, as the value of $t$ increases and the adapted predictor sees more data, as long as the context within the window $[0:T]$ is consistent, the performance of $\adapt[\pred]$ should improve over the pre-trained predictor $\pred$. 
At deployment, $\adapt[\pred]$ should have learned scene context-specific characteristics of human behavior.

\section{Adaptation via Learned Latent Corridors}
\label{sec:methods}
\vspace{-.5em}

\begin{figure*}[t]
  \centering
    \includegraphics[width=0.9\linewidth]{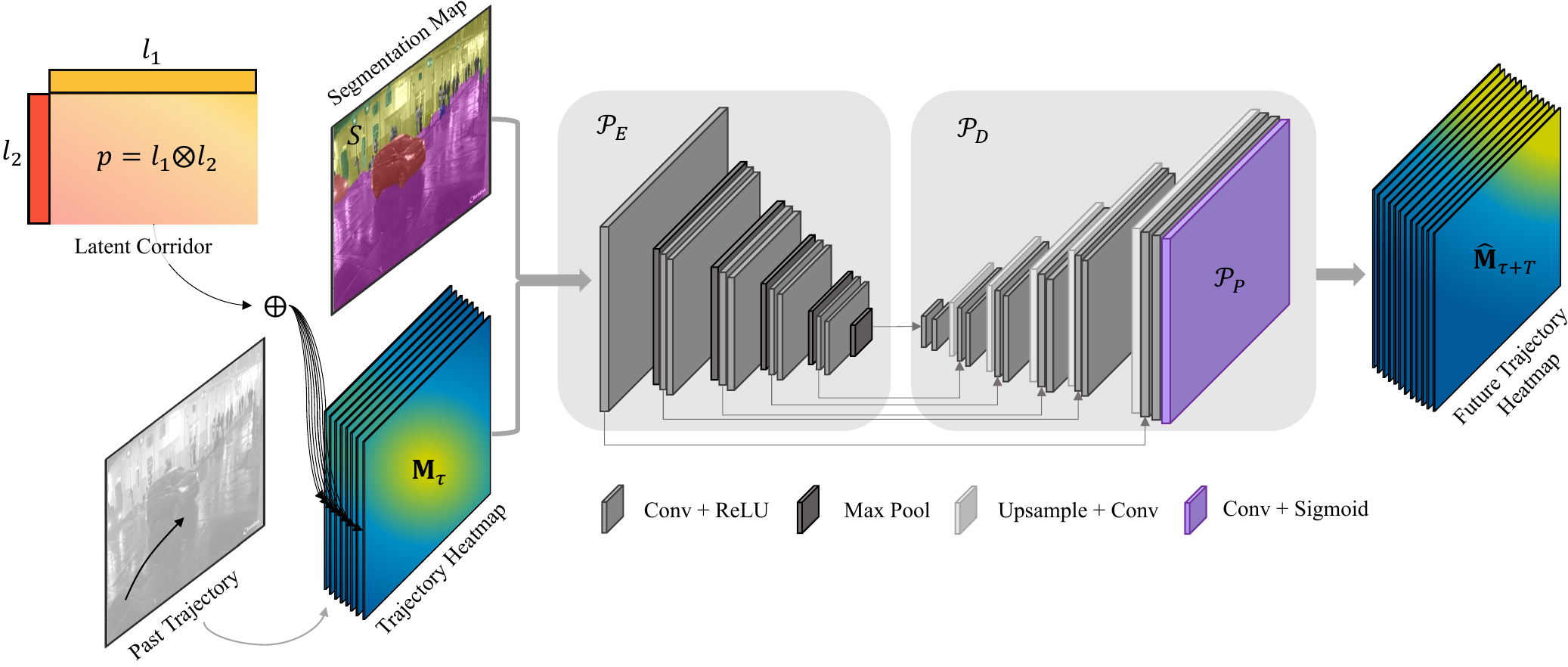}
  \caption{\textbf{Adapting a predictor $\pred$ with latent corridors.} $\pred_E$, $\pred_D$ and $\pred_P$ are pre-trained on the task of human trajectory prediction, taking as input trajectory heatmaps $\heatmap_{\ptime-\histhor:\ptime}$ and segmentation $S$, and outputting predicted trajectory heatmaps $\heatmap_{\ptime+1:\ptime + \predhor}$. We augment $\pred$ with a per-scene \textit{latent corridor} $p$ which is summed element-wise to the input trajectory heatmaps. The latent corridors are trained with $\pred_E$ and $\pred_D$ frozen. The predictor head $\pred_P$ can be frozen, tuned on a single deployment scene, or tuned jointly across multiple scenes.}
  \label{fig:method}
  \vspace{-0.9em}
\end{figure*}
To adapt our trajectory predictor efficiently, we take inspiration from language-based prompt-tuning \cite{lester2021power} and instantiate it within our adaptive trajectory prediction problem. 
Specifically, we augment a frozen base predictor with a learnable latent prompt we call a \textit{latent corridor}, a non-physical corridor that informs the predictor of human behavior patterns in a specific deployment scene. 
In this section, we detail our predictor architecture, prompt representation and training, and investigate a suite of latent corridor prompting configurations. 
% $\pred$ an implicit addition to a predictor $\pred$ that gives $\pred$ a more fine-grained understanding of human behaviour in a specific deployment domain. 
% We implement latent corridors through prompt tuning in image space as described in this section.

\subsection{Base Trajectory Predictor Architecture}
\label{subsec:base_pred}
Our base predictor model, $\pred$, 
% described in Sec.~\ref{subsec:our_atp}, 
consists of three modules: an encoder $\pred_E$, decoder $\pred_D$, and predictor head $\pred_P$ (see Figure~\ref{fig:method}). 
We use the YNet encoder for $\pred_E$, and trajectory decoder architecture for $\pred_D$ and $\pred_P$~\cite{mangalam2021goals}. 

% scene segmentation map $S$ is obtained by passing first frame of the video $I$ through Mask2Former~\cite{cheng2021maskformer, cheng2021mask2former}, a segmentation model, and downsampling the segmentation classes to $12$ semantically meaningful classes for pedestrians in outdoor environments.

\para{Scene and trajectory representation} 
To capture the scene, the RGB image $I$ 
of the first video frame is processed with the Mask2Former semantic segmentation model~\cite{cheng2021maskformer, cheng2021mask2former}. 
The final semantic segmentation map, $S$, is obtained by downsampling the segmentation classes to $C=12$ meaningful classes for pedestrians in outdoor environments.
We follow~\cite{mangalam2021goals}, and convert the history of observed agent positions $x_{\ptime-\histhor:\ptime} \in \mathbb{R}^{\histhor \times 2}$ into a trajectory of heatmaps, $\heatmap_{\ptime-\histhor:\ptime}$, each of the same spatial size as $I$, for a total size of $H \times h \times w$.
% Specifically, an agent's history of positions 
% \abnote{shouldn't this be N x H x 2 for all N agents?} \ntnote{The input the predictor is one trajectory at a time, not all agents trajectories} 
% are converted to heatmaps $\heatmap_{1:H}$ of dimension $H \times h \times w$, where, 
% As in \cite{mangalam2021goals}, the $\heatmap(\sid, i, j) = 2\frac{\lVert (i, j) - x_\sid \rVert}{\max_{(x,y) \in I \lVert (x,y) - x_\sid \rVert}}$, $\sid \in \ptime-\histhor:\ptime$.
Heatmaps are concatenated with the semantic segmentation map $S$; the final input into $\pred$ is $\mathbf{o}_\ptime := ([\heatmap_{\ptime-\histhor:\ptime}, S])$ of size $(C+H) \times h \times w$.
The predictor also outputs a trajectory of heatmaps, $\hat{\heatmap}_{\ptime+1:\ptime+\predhor}$.
The corresponding ground truth $\predhor$-step future agent trajectory, $\mathbf{y}_{\ptime} \in \mathbb{R}^{\predhor \times 2}$, is also converted into a trajectory of heatmaps, $\heatmap_{\ptime+1:\ptime+\predhor}$, for loss computation.  

\para{Loss function} 
% The network is pre-trained using a trajectory prediction loss $\ell$. 
We train the predictor with a binary cross entropy loss on the trajectory heatmaps, 
\begin{equation}
\ell := \sum_{\ptime} \text{BCE}(\pred([\heatmap_{\ptime-\histhor:\ptime}, S]), \heatmap_{\ptime+1:\ptime+\predhor}).
\label{eq:train_loss}
\end{equation} 
For evaluation, a \texttt{softargmax} operation is used to sample 2D points from the predicted heatmap $\hat{\heatmap}_{\ptime+1:\ptime+\predhor}$ as in~\cite{mangalam2021goals}. This yields predictions in x-y pixel space, $\hat{\mathbf{y}}_\ptime = \hat{y}_{\ptime+1:\ptime+\predhor}$, for computing displacement error metrics in Section~\ref{sec:experiments} with respect to 
the ground-truth future positions, $\mathbf{y}_\ptime$.

% \para{Pre-training dataset} 
% % Our dataset $\mathcal{D}$ consists of simulated human pedestrian agents moving around various outdoor environments in 90-second videos captured from a single static RGB camera with any point of view (not necessarily birds eye view). 
% The pre-training data $\mathcal{D}$ consists of simulated human pedestrian agents moving around various outdoor environments in 90-second videos captured from a single static RGB camera with any point of view (not necessarily birds eye view). 
% \abnote{@Neerja should we put more details on MOTsynth data?}

\subsection{Representing \& Learning Latent Corridors}
% \subsection{Latent Corridors: Compact but Spatially Grounded}

Our key idea for adapting trajectory predictors is to augment the frozen base model $\pred$ with a set of trainable prompts, called \textit{latent corridors}, which learn new trends in how humans interact with the scene 
% (e.g., cutting through grass as a shortcut) 
or with each other. 
% (e.g., stopping to watch a street performer). 
% We call these prompts \textit{latent corridors}, since they should capture non-physical passageways that guide human behavior in the scene. 
For effective adaptation, we seek two key properties for our latent corridors: parameter-efficient  (i.e., we want to minimize the number of parameters that are tuned from new deployment data) and spatially scene-grounded (i.e., the latent should have pixel-wise alignment with the scene image). We first outline our prompting approach, and then discuss a compact but spatially-grounded representation of the prompt that is amenable to parameter-efficient learning. 

\para{Latent corridor prompt} For each of the $\snum$ deployment scenes, we introduce a unique trainable prompt, $p_\sid \in \mathbb{R}^{h \times w}$, of the same spatial size as the image $I$ that is input into the predictor (left, Fig.~\ref{fig:method}). 
% The prompt is a continuous matrix of dimension $h \times w$. 
For a network that has adapted to $\snum$ scenes, we have a set $\snum$ of prompts $p := \{p_0, p_1, \ldots, p_K\}$.
Thus, our adaptation rule to scene $k$ is 
\begin{equation}
    \adapt := \heatmap_t \oplus p_k \quad \forall t \in \{\ptime-H, \hdots, \ptime\}.
    \label{eq:promtp_adapt_rule}
\end{equation}
The prompt is summed element-wise to each of the input heatmaps corresponding to the observed trajectories; let $\widetilde{\heatmap} = \heatmap \oplus p$ denote this for any original $\heatmap$.
See Supplement Sec.1 for an ablation on prompt location. 
% See Sec.~\ref{sec:supp-ablation} for an ablation on prompt location. 
The adapted predictor takes as input $\pred([\widetilde{\heatmap}_{\ptime-\histhor:\ptime}, S])$. 

% \para{Latent corridors as image prompts}
\para{A compact but spatially-grounded representation} 
In Equation~\eqref{eq:promtp_adapt_rule}, the prompt is assumed to be the same size as the image, $h \times w$, which is nice for spatial alignment between the prompt and scene. 
However, in our experiments, where all images are of size $h=288$ and $w=480$, this naive image-based representation requires learning an additional $138$K parameters. 
Considering that our base predictor $\pred$ has $\sim900$K trainable parameters, this prompt increases the model parameter size by over $15\%$. 
% and makes it infeasible to learn meaningful latent corridors from small amounts of new human trajectories. 
% Considering that our base predictor $\pred$ has $\sim900$K trainable parameters, this prompt increases the model parameter size by over $15\%$ and makes it infeasible to learn meaningful latent corridors from small amounts of new human trajectories (see Sec.~\ref{sec:supp-ablation}). 
% In our experiments, such as design was unable to learn meaningful latent corridors from small amounts of new human trajectories (see \todo{add reference to supplement?}). 
Instead, we propose a \textit{low-rank representation} of the prompt with rank $1$. We initialize our latent corridor as a vector of dimension $h + w$ using Kaiming initialization~\cite{he2015delving}, and parameterize the full prompt as the outer product of the $h$ and $w$ dimensional vectors. This lightweight representation preserves the spatial relationship between the prompt, scene, and trajectory heatmaps and is significantly more compact: it increases the model parameter size by less than $0.1\%$. We find empirically that this rank 1 matrix performs similarly to a full rank matrix.
% Our base model has over 900K trainable parameters, but for our experiments $h=288$ and $w=480$, so therefore, this lighweight prompting scheme results in an increase of less than $0.1\%$ of the model parameters.

% Given that our base trajectory prediction model does not take transformer tokens, we had to make a non-trivial design choice for the prompt. 
% A latent vector is often a compact, global, representation. 
% Unlike transformer architectures with variable input sizes, our base predictor model has a fixed input size of $(C+H) \times h \times w$. 
% Modifying the input size of model in order to accommodate a latent vector is not possible, therefore, the latent must be added to the model in a way that does not change the input size. Therefore, we choose the latent prompt to be the same size as the input image $I$. 
% This makes the prompt the correct dimension to be summed element-wise to the heatmap input $M$, and allows for the latent to be compact but still correspond to spatial locations in the image. 

\para{Training} We learn the prompt $p$ while the predictor encoder $\pred_E$ and decoder $\pred_D$ are frozen. The predictor head $\pred_P$ is optionally tuned and the latent corridors can be trained individually on one scene at a time, or simultaneously amongst many scenes (see Section~\ref{sec:prompt_configs}). 
We use the same trajectory loss $\ell$ from Equation~\eqref{eq:train_loss} that the base predictor was pre-trained with. 

% We call the trained prompt our latent corridor for a given scene. 

% \subsection{Image Prompt Representation}

\subsection{Prompting configurations}
\label{sec:prompt_configs}

One of the strengths of our latent corridors approach is that it is compatible with a suite of deployment desiderata that can inform how the prompts are incorporated into the predictor. We identify and study three settings. In each setting, $\snum$ unique prompts are trained on $\snum$ deployment scenes individually, but the treatment of the predictor head differs.

\begin{enumerate}
    \item \textbf{Latent corridor adaptation (LC):} The simplest use of latent corridors is to keep the entire base predictor frozen, including the original predictor head $\pred_P$. 
    % \ntnote{I'm not sure if we need the following sentence}Intuitively, by inputting any of the scene-specific latent corridors, the base predictor can adapt to patterns within any of the $\snum$ scenes. 
    % Additionally, the total number of new parameters is $\snum \times (h + w)$, which scales linearly in the number of scenes, making it amenable to more data-scarce regimes. 
    This design is the fastest to train since it has the smallest number of trainable parameters, so it is desirable when rapid adaptation to short-term transient events occurs. It also enables easy recovery of the base predictor model and its original performance by simply not inputting a prompt. 
    
    \item \textbf{Multi-scene finetuning (LC + Joint FT):} In this setting, each of the $\snum$ deployment scenes has a unique latent, but one single predictor head $\pred_P$ is jointly tuned across the $\snum$ scenes. The latents retain unique scene information, but $\pred$ is better adapted to in conjunction with the per-scene latents. Similarly to the LC approach above, this is a more compact configuration since one prediction model is used for all scenes, so it is desirable if deployment hardware has limited space. It is also faster to train than per-scene finetuning especially as $\snum$ grows, since multiple predictor heads do not have to be tuned.

    \item \textbf{Per-scene finetuning (LC + Per-Scene FT):} If one seeks to maximize performance within a \textit{specific} deployment scene, one can \textit{jointly} finetune $\snum$ predictor heads $\pred_{P_k}$ with $\snum$ latent corridors (one per scene). While this results in the need for a unique predictor for each deployment scene, we find empirically that this method achieves best in-scene adaptation performance. 
    % This aligns with intuition that both the latent corridor and the $\pred_P$ parameters are being specialized to one scene. 
\end{enumerate}

% In practice, finetuning the predictor head along with the latent corridors leads to better performance, and so we treat keeping $\pred_P$ frozen as an ablation.

% Perhaps better performance on the new deployment domain is the priority, and finetuning the model makes sense - this could either be done on several scenes jointly, or on individual scenes. 
% Therefore, we investigate our method in the following configurations for the predictor head $\pred_P$:

% \begin{enumerate}
%     \item Frozen $\pred_P$, tune on $\snum$ scenes individually: tune $\snum$ unique prompts
%     \item Tunable $\pred_P$, tune on $\snum$ scenes jointly: tune $\snum$ unique prompts, tune $\pred_P$ jointly on all $\snum$ scenes
%     \item Tunable $\pred_P$, train on $\snum$ scenes individually: train $\snum$ models, each with a unique prompt and unique $\pred_P$
% \end{enumerate} 

% \subsection{Prompting for generalization}

% \todo{Stretch goal - but if we get this to work, describe replacing direct optimization of latent corridor with a learnable module that takes segmentation as input and outputs latent corridor}
\section{Experimental Setup}
\label{sec:experiments}

We study our approach on synthetic and real datasets. Here, we detail these datasets, describe how we evaluate adaptation quality over time, and outline our trajectory predictor baselines. 
Together, our experiments on MOT, WildTrack and EarthCam scenes cover a diverse range of key real-world properties including different lighting conditions, flat vs varied environment topologies, different crowd densities, and a variety of types of scenes.

\subsection{Datasets}
\label{subsec:datasets}
\vspace{-.5em}
\para{MOTSynth} We start in simulation with MOTSynth~\cite{fabbri2021motsynth}, a synthetic pedestrian detection and tracking dataset of over $700$ $90$-second videos with varying camera viewpoints and outdoor environments. Pedestrians carry out simple actions such as walking, standing, or running, and follow manually pre-planned flows as well as a collision avoidance algorithm. MOTSynth has over $17$ hours of video; we select a subset of approximately $11$ hours of video corresponding to the $437$ scenes with a static camera. Due to the large size of the dataset and perfect ground truth detections, MOTSynth was a good starting point. 

\para{MOT \& WildTrack} We also evaluated our approach on the real-world pedestrian datasets MOT and WildTrack. WildTrack~\cite{chavdarova2018wildtrack} consists of $7$ static camera viewpoint videos of pedestrians walking through a plaza. We randomly selected 3 videos and took the first $5$ minutes of each video. MOT, the multiple object tracking benchmark, consists of several video datasets of pedestrians navigating outdoor environments. We combined the datasets MOT15~\cite{leal2015motchallenge}, MOT16~\cite{milan2016mot16}, and MOT20~\cite{dendorfer2020mot20}, and removed videos with dynamic cameras, pedestrian density of less than $10$, and length less than $45$s, leaving 4 scenes: MOT16-03, MOT20-02, AVG-TownCentre, and PETS09-S2L2.

\para{EarthCam} We evaluate our approach in-the-wild on data from \texttt{https://www.\allowbreak earthcam.com}, which has livestream webcam data from around the world available, and from which we scrape four 5 minute segments of data. 

% \begin{itemize}
%     \item MOTSynth~\cite{fabbri2021motsynth}: synthetic pedestrian detection/tracking dataset with static and dynamic scenes, many scenes are very crowded. ~700 videos, ~17 hours of data. Explain that we begin with synthetic data because of the large number of videos with a static camera, perfect ground truth, etc
%     % \item SDD/ETH
%     \item Multiple Object Tracking Benchmark (MOT): contains ~10 datasets for 3D tracking and surveillance, which all have human detections. Describe that we select videos with a static camera that are more than 30 seconds long
%     % \item 
%     % \item VIRAT - surveillance video
%     % \item Multiview Extended Video with Activities dataset - 250+ hours of ground camera video, 12.5 hours are annotated
%     % \item CAVIAR dataset 
%     % \item Surveillance Perspective Human Action Recognition dataset (SPHAR)
%     \item WildTrack dataset 
%     \item ETH/UCY
% \end{itemize}

\subsection{Train-Test Split Over Time in Human Seconds}
\label{subsec:train-test}
Given a video of pedestrians in motion, we prepare our dataset to adapt to the scene as follows. 
We use provided annotations or run ByteTrack~\cite{zhang2022bytetrack} to get tracklets of all $\nagents$ detectable identities in the video. Tracklets are downsampled to 2 timesteps per second, then windowed by $20$ timesteps with an observed length $H=8$ and future length $T=12$. Then, we sort the $\nagents$ identities chronologically by their first appearance in the scene. We select the first $80\%$ of identities that appear in the scene for our training dataset, and hold out the last $20\%$ for the testing set. When we conduct experiments over time, the testing set is always the same $20\%$ of identities, whereas the training set consists of the first $m\%$ of identities, $m \le 80\%$. We report adaptation time in human-seconds of observation, where 1 person for 1 second is 1 human-second, so for instance, 30 people observed for 10 seconds each would be 300 human-seconds.

\subsection{Base Predictor Pre-Training Dataset}
\label{subsec:base_pred_training}

We first pre-train our base predictor $\pred$ described in Sec~\ref{subsec:base_pred} on a dataset $\mathcal{D}$ which consists of simulated human pedestrian agents moving around various outdoor environments, captured from a single static RGB camera with any point of view (not necessarily birds eye view). This dataset comes from 437 90-second MOTSynth videos with a static camera. We pre-train our predictor on a train-test split over time of $\mathcal{D}$ as described in the previous section.

\subsection{Baselines}
\label{subsec:baselines}
We evaluate our method with the following baselines.

\begin{enumerate}
    \item \textit{Constant velocity}: As in~\cite{599328}. 
    \item \textit{Learned trajectory}: Simplified PECNet~\cite{mangalam2020not}, which is a predictor learned from position histories but no scene information. 
    % but learn a more sophisticated model of how humans move from trajectories alone. 
    
    \item \textit{Scene-aware}: Simplified YNet~\cite{mangalam2021goals}, which is a learned predictor with position and scene input. See Supplement Sec. 3 for a comparison with YNet.
    \item \textit{ATP Finetune}: Adaptation baseline which finetunes scene-aware predictor head $\pred_P$ without latent corridors. 
    % Only predictor head $\pred_P$ of the scene-aware baseline is finetuned. 
    % Note that for configuration 2 this baseline is not applicable if the $K$ scenes being jointly tuned on are the same as the pre-training set.
\end{enumerate}

\vspace{-.5em}We implement the learned trajectory baseline by taking the encoder, decoder and trajectory loss from PECNet~\cite{mangalam2020not}; we use a successful architecture on the trajectory prediction problem, removing multimodality for simplicity. Similarly, YNet's~\cite{mangalam2021goals} encoder and trajectory decoder are used for the scene-aware baseline. 
% We evaluate our method in the multimodal setting with PECNet and YNet off-the-shelf as baselines.

\vspace{-.5em}
\subsection{Metrics}
\vspace{-.5em}

Our experiments are evaluated using the Average Displacement Error (ADE)~\cite{pellegrini2009you} and Final Displacement Error (FDE)~\cite{alahi2016social} metrics. ADE is the average $l_2$ error between the entire predicted and ground truth trajectory, while FDE is the $l_2$ error between just the final point of the predicted and ground truth trajectories.

\section{Results}
\label{sec:experiments}
\vspace{-0.5em}
Here, we detail quantitative and qualitative results on MOTSynth, MOT, and WildTrack, and EarthCam datasets.  

\begin{figure*}[t]
  \centering
    \includegraphics[width=0.9\textwidth]{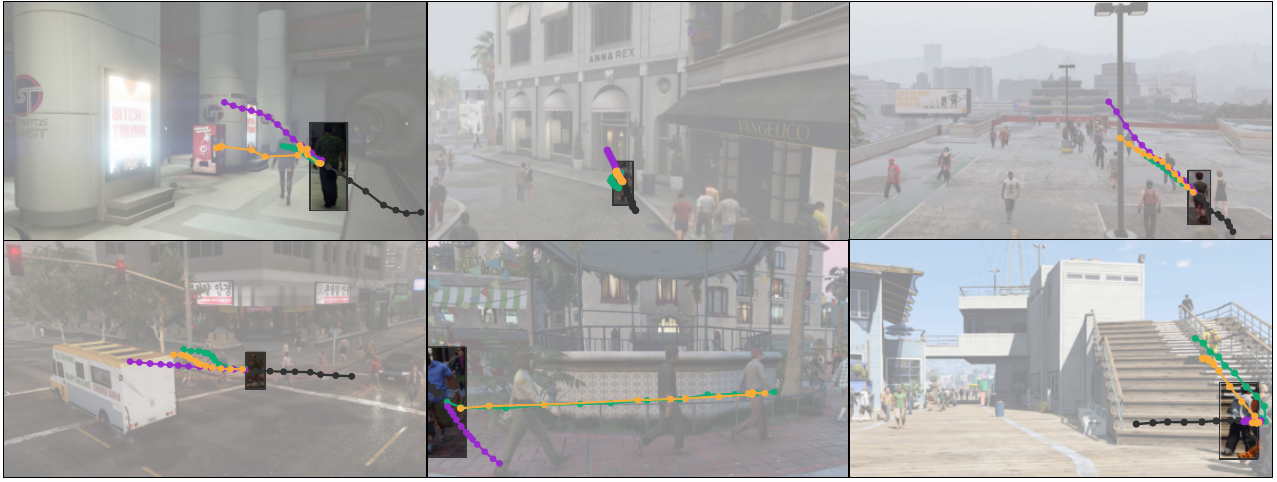}
    \includegraphics[width=0.9\textwidth]{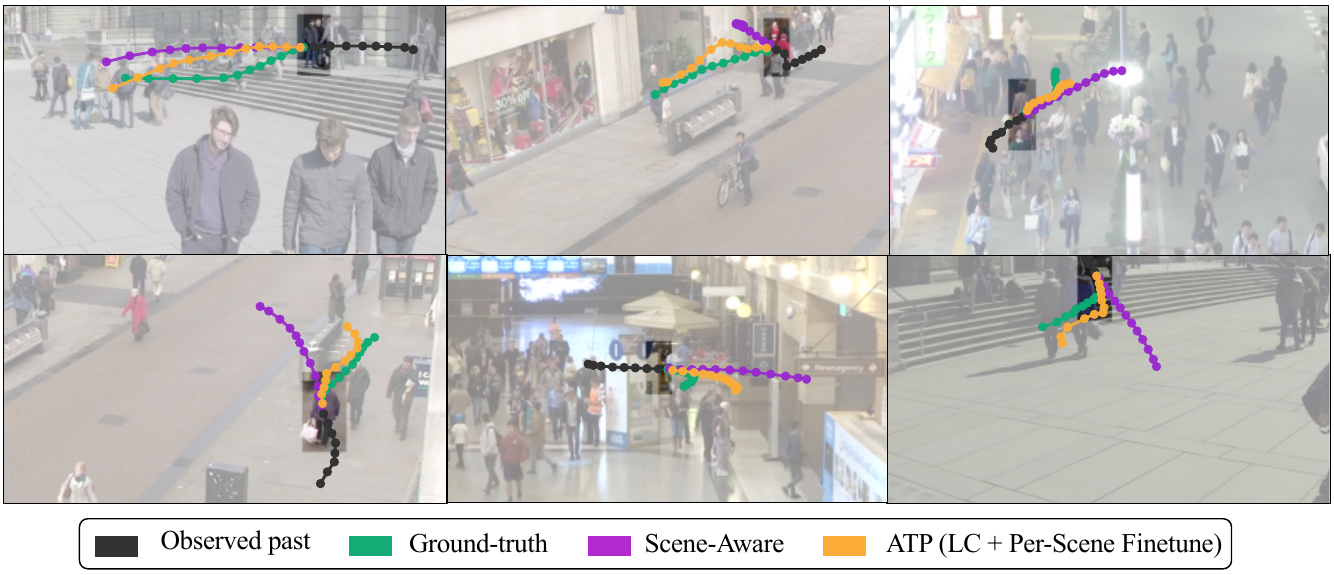}
     % \fbox{\rule{0pt}{2in} \rule{.9\linewidth}{0pt}}
  \caption{\textbf{Qualitative results} on MotSynth (top; synthetic) and MOT and WildTrack (bottom; real). These examples show scenarios where our LC + per-scene finetune ATP method (orange) outperforms the scene-aware baseline (purple). In several MOTSynth examples, the baseline predicts the pedestrian floats into the air (top row), while our method has gained awareness of where the 3D ground plane lies in the 2D image. We also note that patterns of behaviour such as walking on the sidewalk instead of into the road (second row left) and walking up the traversable portion of stairs (second row right) are captured.  On real data, we observe similar awareness of the ground plane and obstacles, as well as a better understanding of nuanced human behavior patterns such as crossing diagonally across a plaza.}
  \label{fig:qual_results}
  \vspace{-1.5em}
\end{figure*}

\begin{figure}[t]
  \centering
    \includegraphics[width=0.99\linewidth]{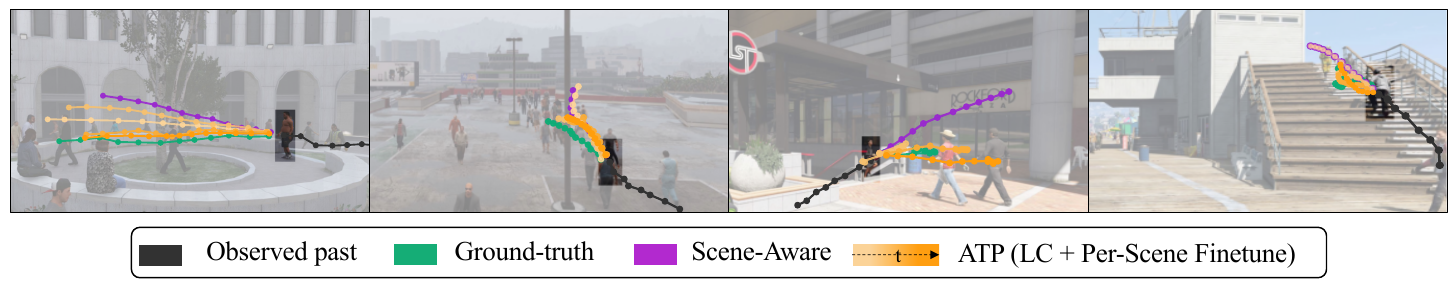}
     % \fbox{\rule{0pt}{2in} \rule{.9\linewidth}{0pt}}
  \caption{\textbf{Qualitative results over time.} Our method's predictions trained on a short number of human seconds ($2\%$) are shown in light orange, to dark orange for a human seconds training time of $80\%$. With the latent corridor trained on a tiny amount of data, the predictions can significantly improve, but at times are close to the baseline. When more human seconds of data are seen, the adaptation results consistently improve. }
  \label{fig:qual_results_over_time}
  \vspace{-0.8em}
\end{figure}

\begin{figure*}[t]
  \centering
     \includegraphics[width=0.95\linewidth]{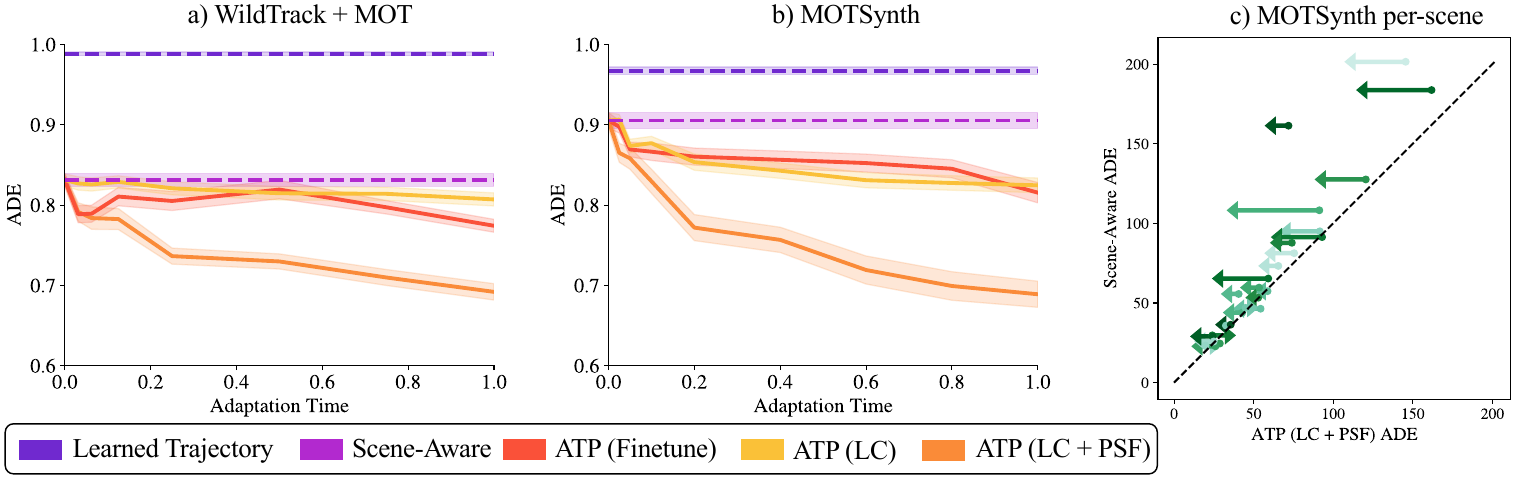}
  \caption{\textbf{Learning latent corridors over time.} (a and b): The x-axis represents adaptation time in human-seconds, or the amount trajectories used to train $\adapt(\pred)$, and the y-axis represents the ADE. Results are normalized per-scene and averaged over models trained on 25 MOTSynth scenes (a) and 7 from MOT and WildTrack (b), with shaded area $\sigma/10$. On all data, even with a short adaptation time, our methods improve on the baselines, and as adaptation time increases, performance improves. Latent corridors + per-scene finetuning has the best performance. c) Comparison to baseline over many MOTSynth scenes for models trained with $8\%$ (point) and $80\%$ (arrowhead) human-second datasets. Each arrow represents one scene, with the ADE using our ATP method plotted against the scene-aware baseline ADE. For some deployment scenes, the scene-aware baseline suffices, whereas other scenes see much more significant benefits from our method.}
  \label{fig:lifelong_MS}
  \vspace{-1.5em}
\end{figure*}

\begin{table*}[t]
\centering
\begin{tabular}{l|ll|ll|ll|ll|ll}
% \textbf{Method} & \multicolumn{2}{c|}{\textbf{MOTSynth}} & \multicolumn{2}{c|}{\textbf{MOT}} & \multicolumn{2}{c|}{\textbf{Wildtrack}} & \multicolumn{2}{c|}{\textbf{NoLA}} \\
%                 & ADE & FDE    & ADE  & FDE    & ADE & FDE    & ADE  & FDE   \\ \hline
% Constant velocity  & 78.5 & 160.4  & 47.7& 99.3   & 44.9 & 90.1   & 42.4  & 82.8  \\
% Learned trajectory (PECNet-Ours) & 51.2 & 100.0  & 49.7   & 103.4  & 43.8  & 83.1   & 36.9  & 65.9  \\
% Scene-aware (YNet-Ours) & 47.3 & 96.5   & 44.2 & 91.0   & 33.6  & 67.7   & 34.0  & 65.1  \\
% ATP (Finetune)  &  - & - & 41.8  & 86.9   & 31.7  & 63.5  & 32.6& 59.6 \\
% ATP (LC) & 44.6 & 90.2   & 43.0 & 89.2   & 32.9  & 65.9   & 33.5 & 63.1  \\
% ATP (LC + Joint Finetune) & \textbf{42.6} & \textbf{85.3} & - & - & - &- & - & - \\
% ATP (LC + Per-Scene Finetune) & - &- & \textbf{37.4}  & \textbf{74.3} & \textbf{27.4}  & \textbf{54.7} & \textbf{26.1} & \textbf{47.4} \\
% \end{tabular}
\textbf{Method} & \multicolumn{2}{c|}{\textbf{MOTSynth}} & \multicolumn{2}{c|}{\textbf{MOT}} & \multicolumn{2}{c|}{\textbf{Wildtrack}} & \multicolumn{2}{c|}{\textbf{EarthCam}} \\
                & ADE & FDE    & ADE  & FDE    & ADE & FDE    & ADE  & FDE   \\ \hline
Constant velocity  & 78.5 & 160.4  & 47.7& 99.3   & 44.9 & 90.1   & 27.3 &52.2  \\
Learned trajectory (PECNet-Ours) & 51.2 & 100.0  & 49.7   & 103.4  & 43.8  & 83.1   & 34.3  & 55.5  \\
Scene-aware (YNet-Ours) & 47.3 & 96.5   & 44.2 & 91.0   & 33.6  & 67.7   & 23.9  & 44.4  \\
ATP (Finetune)  &  - & - & 41.8  & 86.9   & 31.7  & 63.5  & 21.9 & 38.7 \\
ATP (LC) & 44.6 & 90.2   & 43.0 & 89.2   & 32.9  & 65.9   & 23.4 & 42.8  \\
ATP (LC + Joint Finetune) & \textbf{42.6} & \textbf{85.3} & - & - & - &- & - & - \\
ATP (LC + Per-Scene Finetune) & - &- & \textbf{37.4}  & \textbf{74.3} & \textbf{27.4}  & \textbf{54.7} & \textbf{17.5} & \textbf{30.8} \\
\end{tabular}

\caption{\textbf{Our method vs baselines on synthetic and real-world datasets.} The average ADE and FDE in pixel space of (left-to-right) 437 MOTSynth scenes, 4 MOT scenes, 3 WildTrack scenes, and 4 EarthCam scenes. Across these synthetic and real deployment scenes, our latent corridor-only adaptation method is comparable to finetuning the predictor head and outperforms non-adaptive baselines. Our method with latent corridors and per-scene finetuning consistently outperforms all other approaches.}
\label{tab:mot_results}
\vspace{-1.8em}
\end{table*}

                % \begin{table}[H]
                % \centering
                % \begin{tabular}{|l|c|c|}
                % \hline
                % Model & ADE & FDE \\ \hline
                % linear & 47.7 & 99.28 \\
                % blind AE & 46.8 & 96.9 \\
                % ynet & 39.4 & 82.38 \\
                % ft only & 70.02 & 30.7 \\
                % prmpt no ft & 37.9 & 80.42 \\
                % prmpt with ft & 34.25 & 66.42 \\
                % \hline
                % \end{tabular}
                % \caption{Results for the Mean dataset.}
                % \label{tab:results_mean}
                % \end{table}

% \begin{figure}[t]
%   \centering
%      \includegraphics[width=0.99\linewidth]{fig/original_ADE_vs_best_ADE_data_percent_0.8_vs_0.08_finetune and prompting.png}
%   \caption{Lifelong Learning on MOTSynth: Comparison to baseline over many scenes for $t=0.08$ (dots) and $t=0.8$ (X). Each point represents one scene from MOTSynth, with the ADE using ATP for two different adaptation times plotted against the ADE for the scene-aware baseline. Even at a short time horizon, our approach yields significant performance gains over the baseline on many scene. For some deployment scenes, the scene-aware baseline works well, whereas other scenes see much more significant benefits from our method.}
%   \label{fig:lifelong_MS_comparison}
% \end{figure}

\subsection{MOTSynth Results}
\vspace{-0.5em}
We first evaluate predictor improvement enabled by latent corridors when the deployment scene is in the pre-training dataset, $\mathcal{D}$. 
% In order to understand the performance 
% of our method 
% over a large dataset 
% of our approach and the baselines to identical large datasets, 
We train the baselines and two variants of our ATP method (LC and LC + Joint Finetune, described in Sec.~\ref{sec:prompt_configs}) on the MOTSynth pre-training dataset $\mathcal{D}$. 
All models see the same training data, and both the scene-aware baseline and our adaptive method are conditioned on scene semantic segmentation maps.
Results are in the first column of Table~\ref{tab:mot_results}. 
The learned trajectory baseline significantly improves over constant velocity, and adding in scene awareness results in a $7.6\%$ performance gain on ADE and $3.5\%$ gain on FDE. 
Our latent corridor approach results in a $5.7\%$ and $9.9\%$ improvement over the scene-aware baseline for ADE, without and with joint finetuning of the predictor layer respectively, and a $6.5\%$ and $11.6\%$ improvement on FDE.  
This indicates that our latent corridor approach more effectively learns scene-conditioned information that is useful for the trajectory prediction task.

% \para{Prompting configurations} 
Next, for a random subset of $25$ of the MOTSynth scenes, we additionally train latent corridors with per-scene finetuning (LC + Per-Scene FT, Sec~\ref{sec:prompt_configs}). 
The datasets $\mathcal{D}'_{0:t}$ correspond to increasing human-second lengths. 
The test set trajectories are the last $20\%$ of agents in the deployment scene, and the training sets consist of the first $2, 4, 8, 16, 32, 48, 64$ and $80\%$ of agents to enter the scene. Results normalized per-scene and averaged are shown in Fig.~\ref{fig:lifelong_MS}b. Using only latent corridors performs comparably to only finetuning the last layer for each deployment scene, while latent corridors with per-scene finetuning yields significant performance gains. 
% Even short time horizons wherein a small number of agents are observed enables our adaptive predictor to outperform all non-adaptive baselines. 
% the performance gains on the non-adaptive baselines are significant across all cases.
We visualize the ADE results per-scene with our LC + finetuning method trained with $8\%$ and then trained with $80\%$ of the data  in Fig.~\ref{fig:lifelong_MS}c. 
We see that while the effectiveness of the adaptation for varying time horizons differs for each deployment scene, there are many scenes where adaptation yields significant gains, and some where the improvement on ADE error is up to $63.4\%$. 
We hypothesize that scenes where our method yields smaller gains exhibit behaviors and environment geometries that the prior $\pred$ is sufficient for.

\vspace{-.5em}

\para{Qualitative results} Visualizations from the per-scene models can be seen in Fig.~\ref{fig:qual_results} and Fig.~\ref{fig:qual_results_over_time}. 
% Since the pedestrian motion in this dataset follows a simple simulated model, the behaviours are relatively contrived and unrealistic. Therefore, we do not evaluate the degree to which human behaviours seem to be captured with our approach on synthetic data. 
We observe many examples where the scene-aware baseline seems to lack awareness of the ground plane. 
In multiple instances, the scene-aware baseline predicts a pedestrian floats into the air (Fig.~\ref{fig:qual_results} top row and Fig.~\ref{fig:qual_results_over_time}), whereas our predictions lie closely in line with the ground truth trajectory in terms of distance from the ground plane. This holds even for non-planar ground planes: When a pedestrian walks up stairs, our model seems to better understand their 3D structure, for example in the bottom right of Fig.~\ref{fig:qual_results_over_time}. Additionally, our approach captures trends of behaviour in scenes such as walking on the sidewalk instead of into the road, walking around the rotunda, and walking up the section of smaller, easily traversable steps (see Fig.~\ref{fig:qual_results} second row, left to right).

\subsection{MOT and WildTrack Results}
\label{subsec:MOT}

We next investigate if latent corridors help predictor adaptation when the deployment scene is \textit{outside} of the pre-training dataset, and if latent corridors outperform ATP via direct finetuning on deployment data.
Specifically, we use the base predictor $\pred$ which only saw MOTSynth data $\mathcal{D}$ and then 
directly do sim-to-real adaptation via latent prompting, finetuning, or a combination thereof 
% then learn latent corridors + per-scene finetuning 
using each of seven \textit{real human pedestrian scenes} from MOT and Wildtrack as $\mathcal{D}'$.
% The  predictor $\pred$ was pre-trained on synthetic scenes scenes, and we directly do sim-to-real via latent prompting, finetuning, or a combination thereof. 
Plots of the ADE over time averaged over these scenes is shown in Figure~\ref{fig:lifelong_MS}a. 
Regardless of the ATP setting, adaptation over time results in a consistent error reduction compared to the baselines. 
While only finetuning seems to be slightly more effective on real-world scenes than prompting alone, our latent corridors + per-scene finetuning approach is significantly more effective ($11.2\%$) than finetuning alone (see Table~\ref{tab:mot_results}).

\para{Qualitative results} Visualizations of baselines and our LC + PSF approach on MOT and WildTrack are in Fig.~\ref{fig:qual_results}. Similar trends from experiments on synthetic data carry over. Our approach enables the predictor to better ground future behavior in the scene geometry: for example, pedestrians are no longer predicted to float upwards (third row, left and middle), and future behavior is guided by a finer-grained awareness of obstacles (third row right, bottom row left and middle). 
We also observe our adapted predictor learning trends in human behavior: in the WildTrack plaza environment shown in Fig.~\ref{fig:teaser} middle row and Fig.~\ref{fig:qual_results} bottom right, pedestrians tend to avoid the middle of the plaza and instead cross it diagonally. Our method learns this subtle, scene-specific behavior pattern and thus predicts more accurately.

% \subsection{Multimodal Results on MOTSynth}
% \begin{enumerate}
%     \item Similar table to table~\ref{tab:motsynth_results} but for K=1, K=5 and K=20, results on full YNet
% \end{enumerate}

\subsection{EarthCam Results}

\begin{figure}[t]
  \centering
     \includegraphics[width=0.99\linewidth]{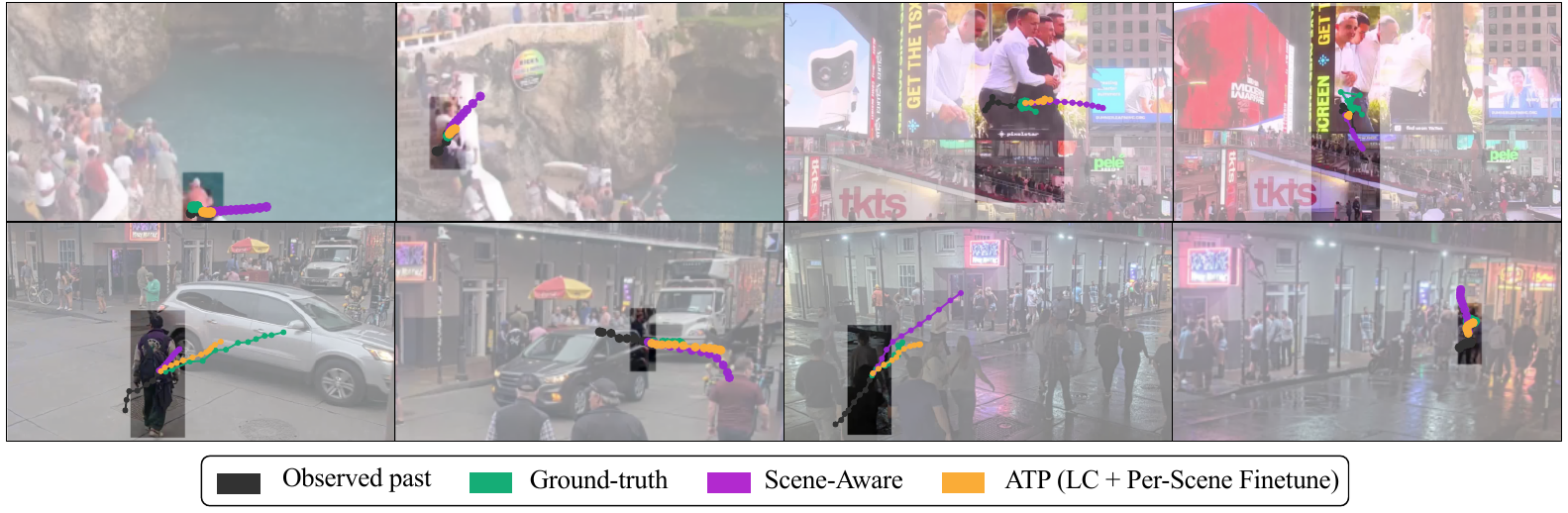}

  \caption{\textbf{Results on EarthCam webcam data.} At a cafe in Jamaica (top left 2 panels) with an overlook and stairs on the edge of the water, our model (orange) is able to correctly predict that people will stick to moving up and down the path created by the stairs, while the scene-aware baseline (purple) predicts people will walk directly into the water or over the edge of the path. In Times Square, a billboard depicts human actors in motion (top right 2 panels). The scene-aware baseline assumes that these actors will move following their observed history, while our model correctly predicts that anyone in the area of the billboard will stay in the billboard.  We train our method on one NoLA intersection during the daytime (bottom left 2 panels), when pedestrians must obey vehicles moving through the street, and at nighttime (bottom right 2 panels), when pedestrians take over. During the day, the model learns to account for humans walking through crosswalks instead of diagonally into the intersection, and to account for transient push-carts in the area.}
  \label{fig:earthcam}
  \vspace{-1.5em}
\end{figure}

\vspace{-0.5em}

Finally, we evaluate the performance of our approach on truly ``in-the-wild'' data by scraping four 5-minute videos of pedestrian data captured in different locations: Rick's Cafe in Jamaica, Times Square in New York City, and Bourbon Street in New Orleans during daytime and nighttime. We use ATP as in Sec.~\ref{subsec:MOT}.
% activity (top row, Fig.~\ref{fig:nola}) and one video during nighttime activity (bottom row, Fig.~\ref{fig:nola}). 
% We scraped two additional 5-minute videos from EarthCam as described in Sec.~\ref{subsec:datasets}, one from Rick's Cafe in Jamaica and one from Times Square in New York City. Qualitative results for these scenes are in Fig.~\ref{fig:webcam-qual-supp}. In the top row, 
% Additional results on different scenes of webcam data are in Sec~\ref{sec:supp-webcam}.
Quantitatively, the results on this in-the-wild data align with our results on real and synthetic data: our ATP model with latent corridors and per-scene finetuning outperforms the non-adaptive baselines and pure finetuning (see Table~\ref{tab:mot_results}).
Qualitatively in Fig.~\ref{fig:earthcam}, we see a variety of interesting adaptations. At the cafe, there is a complicated path through an overlook that twists down stairs towards the water. The scene-aware baseline often predicts that people will jump over a ledge, whereas our method learns the boundaries of the paths that people follow and is able to predict that people will stay within those boundaries. In Times Square, the scene-aware baseline does not recognize that people on a billboard will stay within the billboard, whereas our method is able to recognize that.  In the NoLA videos, our method adapts to daytime patterns where pedestrians navigate around carts and don't frequently walk diagonally through streets because of through-traffic, but when the adaptive predictor observes nighttime behavior, it no longer has to respect these daytime patterns and learns to predict pedestrians as crossing diagonally.
Similarly to the prior datasets, our ATP model again learns to ground pedestrian future behavior in the 3D ground plane.

\vspace{-0.5em}
% \subsection{Generalization Results on MOTSynth}
% \begin{enumerate}
%     \item Stretch goal - if we get this working
% \end{enumerate}
% \red{\section{Discussion: Offline Adaptation}
% We believe ATP is a problem of great practical utility when deploying systems in the real world with changing conditions and environments. As opposed to envisioning our method being used in real time, we focus on the setting of \textit{offline} adaptation. Imagine a setting with several surveillance cameras set up in a downtown business district. The flow of pedestrians during the rush hour before work on the weekday will be distinct from that on a weekend morning. Additionally, every camera will capture a distinct 3D environment. During the course of a week, latent corridors could be trained to adapt to the varying scene and context specific behaviours present on different days/times. Once training is complete, adapted models could be used \textit{online} for several months, until perhaps a change of season causes a significant change in behaviour. Since our method requires less than $0.1\%$ additional parameters per adaptation context, swapping different latent corridors for different times of day or locations is lightweight, even on device. An online version of our method or others within our proposed ATP framework would be an exciting avenue for future research.}
\vspace{-.5em}
\section{Conclusion}
\label{sec:discussion}
\vspace{-0.5em}
In this work, we formalize and study the problem of adaptive trajectory prediction: the ability of human predictors to adapt to changing deployment conditions and environments. 
To this end, we proposed a lightweight adaptation approach grounded in image-based prompt tuning called latent corridors.  Through extensive experiments on both simulated and real-world pedestrian datasets, we observed that latent corridors enable a data-efficient way to adapt pre-trained predictors to new deployment-scene-specific human behavior. 
% Future work \abnote{put in one sentence on future work?}
% We hope that this work will encourage more research in adaptive trajectory prediction.

% Future directions:
% \begin{enumerate}
%     \item Generalization to new scenes: use meta-learned latent corridors to infer the latent corridors from a new scene image + a very small amount of trajectories (perhaps 1 second or 1-5 frames), using both general knowledge and quickly adapting to specialized knowledge from a scene (note: we should probably try to do this in time for rebuttal or resubmission to ECCV)
%     \item Extract 3D information from latent such as depth map
%     \item Get method to work on dynamic scenes with egocentric video - likely would require re-structuring latent. Then, a robot navigating a scene with people could use this method to improve after 10-15 seconds of navigation
%     \item Bake in more priors (ex. where people are facing), do this in 3D
% \end{enumerate}
\paragraph{Acknowledgements} 
We thank Anastasios Angelopoulos, Antonio Loquercio, and Jathushan Rajasegaran for useful discussions and feedback. This work was supported by ONR MURI N00014-21-1-2801 and a NSF Graduate Fellowship.

\bibliographystyle{splncs04}
\bibliography{main}

\title{Supplementary Material: Adaptive Human Trajectory Prediction via Latent Corridors} 

\titlerunning{Supplementary Material: ATP via Latent Corridors}

% Second Author\inst{2,3}\orcidlink{1111-2222-3333-4444} \and
% Third Author\inst{3}\orcidlink{2222--3333-4444-5555}}

% % TODO FINAL: Replace with an abbreviated list of authors.

% % TODO FINAL: Replace with your institution list.
% Springer Heidelberg, Tiergartenstr.~17, 69121 Heidelberg, Germany

\author{Neerja Thakkar\inst{1} \and Karttikeya Mangalam\inst{1} \and Andrea Bajcsy\inst{2} \and Jitendra Malik\inst{1}}

\authorrunning{N. Thakkar et al.}

\institute{UC Berkeley \and Carnegie Mellon University}

\maketitle

% \clearpage
% \setcounter{page}{1}
% \maketitlesupplementary

In the supplementary material, we provide an ablation on the implementation of our prompting method and a visual overview of our ATP problem formulation. We also show additional qualitative results on MOTSynth and more detailed quantitative results on EarthCam data, quantitative results on ETH/UCY, and compare our implementation of the scene-aware baseline to the original YNet paper~\cite{mangalam2021goals}. Video results can be seen at \href{https://neerja.me/atp_latent_corridors/}{this webpage}.

\section{Details on Latent Corridors for Adaptive Trajectory Prediction}
\label{sec:supp-ablation}

\subsection{Adaptive Trajectory Prediction Visualization}

\begin{figure}[t]
  \centering
    \includegraphics[width=0.99\linewidth]{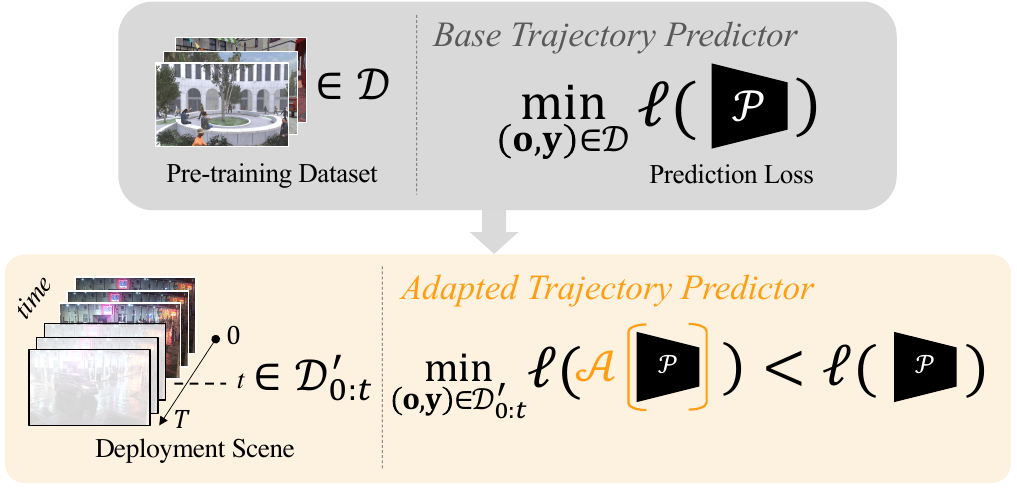}
  \caption{\textbf{Adaptive trajectory prediction}. ATP, formulated in Sec. 3, allows a pre-trained predictor $\pred$ to adapt to a new deployment scene by learning over time on the deployment scene. Once adaptation has occurred, the adapted predictor $\adapt[\pred]$ should perform better on the deployment scene. }
  \label{fig:ATP}
\end{figure}

We provide an illustrative overview of adaptive trajectory prediction problem, formulated in main text Sec. 3, in Fig.~\ref{fig:ATP}.

\subsection{ATP via Latent Corridors on Architectures Beyond YNet}

ATP is an architecture-agnostic paradigm, and latent corridors are also not specific to Y-Net but rather can also work on different architectures. To demonstrate this, we experimented with the Learned Trajectory (PECNet-Ours) architecture on the 473 MOTSynth scenes. Taking the pretrained PECNet-Ours model, we summed a per-scene 16 parameter latent directly to the input, eight $xy$ coordinates. Training the 16D latent along with finetuning the last layer of PECNet-Ours, we see an improvement of $10.2\%$ on ADE and $10.4\%$ on FDE.

\begin{table}[t]
\centering
\begin{tabular}{lll}
Method                 & ADE           & FDE      \\ \hline
Learned Trajectory (PECNet-Ours)      &  51.2 & 100.0  \\
ATP (LC + Joint Finetune) & \textbf{46.0}&  \textbf{89.6}\\
\end{tabular}
\caption{Results of applying ATP via latent corridors to PECNet. ATP using 16D latent corridors with joint finetuning improves the performance of PECNet.}
\label{tab:pecnet_latents}
\end{table}

\subsection{Ablation on Prompt Location/Size}

We ablated the prompt location and method of combining with the input. For the input location, we experimented with combining the prompt with several parts of the input to $\pred$, $[\heatmap_{\ptime-\histhor:\ptime}, S]$: all of the input heatmaps $\heatmap_{\ptime-\histhor:\ptime}$, just the first input heatmap $\heatmap_0$, just the segmentation map $S$, and all of the inputs. For method of combination, we experimented with element-wise summing and element-wise multiplication. We ran this ablation on the latent corridors-only approach to training on MOTSynth. Results can be seen in Table~\ref{tab:prompt_ablation}. While summing seems to lead to better performance than multiplying, and summing to the heatmaps seems to be more helpful than summing to the segmentation, generally, the prompts are effective at improving performance on a variety of input locations. This shows promise for training latent corridors to adapt a variety of architectures.

\begin{table}[t]
\centering
\begin{tabular}{lll}
Prompt Method                 & $ADE$           & $FDE$       \\ \hline
Sum to all heatmaps $\heatmap_{\ptime-\histhor:\ptime}$                  &  \textbf{44.6} &\textbf{90.2}  \\
Sum to  $\heatmap_0$   & 45.1 & 92.1 \\
Sum to  $S$ & 46.4&  97.3\\
Sum to $\heatmap_{\ptime-\histhor:\ptime}$ and $S$ &   46.4 & 96.5\\
Multiply to all heatmaps $\heatmap_{\ptime-\histhor:\ptime}$                  &  46.5 &96.1  \\
Multiply to  $S$ & 45.8&  93.6\\
Multiply to  $\heatmap_0$   & 47.6 & 101.6 \\
Multiply to $\heatmap_{\ptime-\histhor:\ptime}$ and $S$ &   46.5 & 96.1\\
\end{tabular}
\caption{Ablation on 437 MOTSynth scenes in the ATP latent corridor adaptation configuration. We experiment with different locations and methods of combining the prompt with the input, and find that summing the prompt to all input heatmaps yields the best result, but most combinations result in an improvement on the scene-aware baseline (see main text Table 1).}
\label{tab:prompt_ablation}
\end{table}

\begin{figure*}[t]
  \centering
    \includegraphics[width=0.99\linewidth]{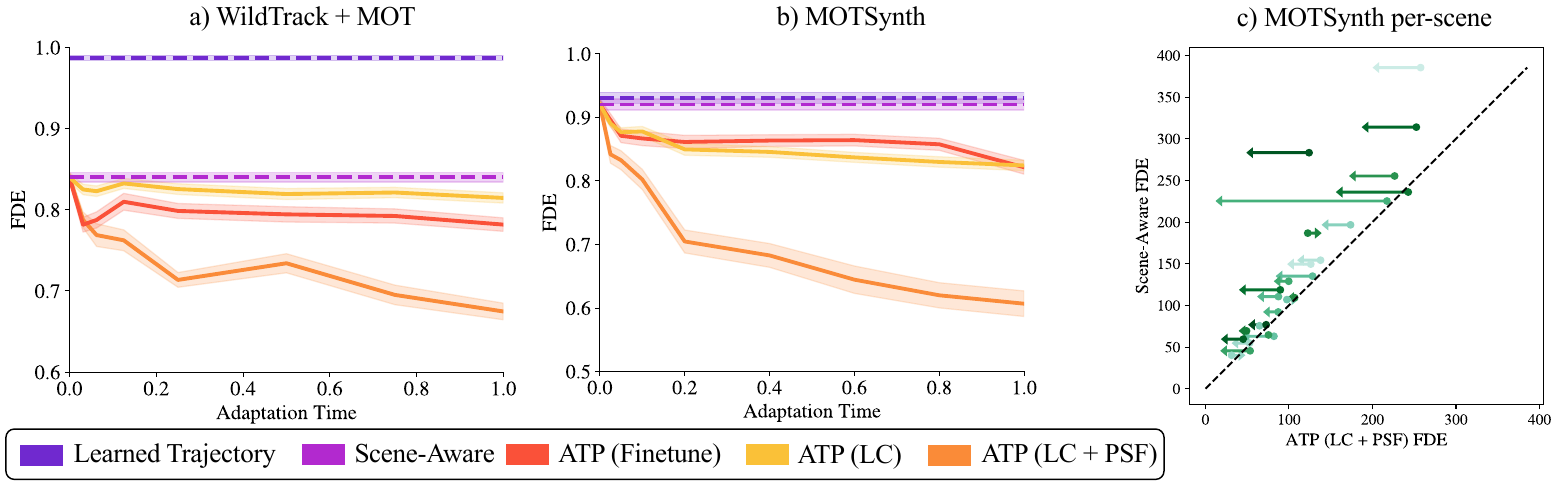}
  \caption{\textbf{Adaptation over time FDE}. As in main text Fig. 5, the x-axis represents normalized adaptation time in human-seconds. The y-axis represents the FDE (lower is better). Results are normalized per-scene and averaged over models trained on 25 MOTSynth scenes (a) and 7 from MOT and WildTrack (b), with shaded area $\sigma/10$. For the FDE metric, our methods improve on the baselines increasingly with adaptation time. Latent corridors + per-scene finetuning has the best performance, as with FDE, and ATP via just finetuning or just latent corridor learning is still comparable. c) Comparison to baseline over many MOTSynth scenes for models trained with $8\%$ (point) and $80\%$ (arrowhead) human-second datasets for FDE. For many deployment scenes, FDE improves significantly more with our method than ADE improved, but still, the per-scene improvements are varied. }
  \label{fig:adaptation_time_FDE}
\end{figure*}

\subsection{Segmentation Classes}

We condense Mask2Former’s 132 classes into 12 classes that are meaningful for outdoor pedestrians: person,  bicycle, car, motorcycle, large vehicle, traffic light, stop sign, bench, stairs, road/ground, building/wall, other. 

\section{Additional Experimental Results}
\label{sec:supp-webcam}

\subsection{Results on MOTSynth, WildTrack, MOT, and EarthCam}

\begin{figure*}[t]
  \centering
    \includegraphics[width=0.95\linewidth]{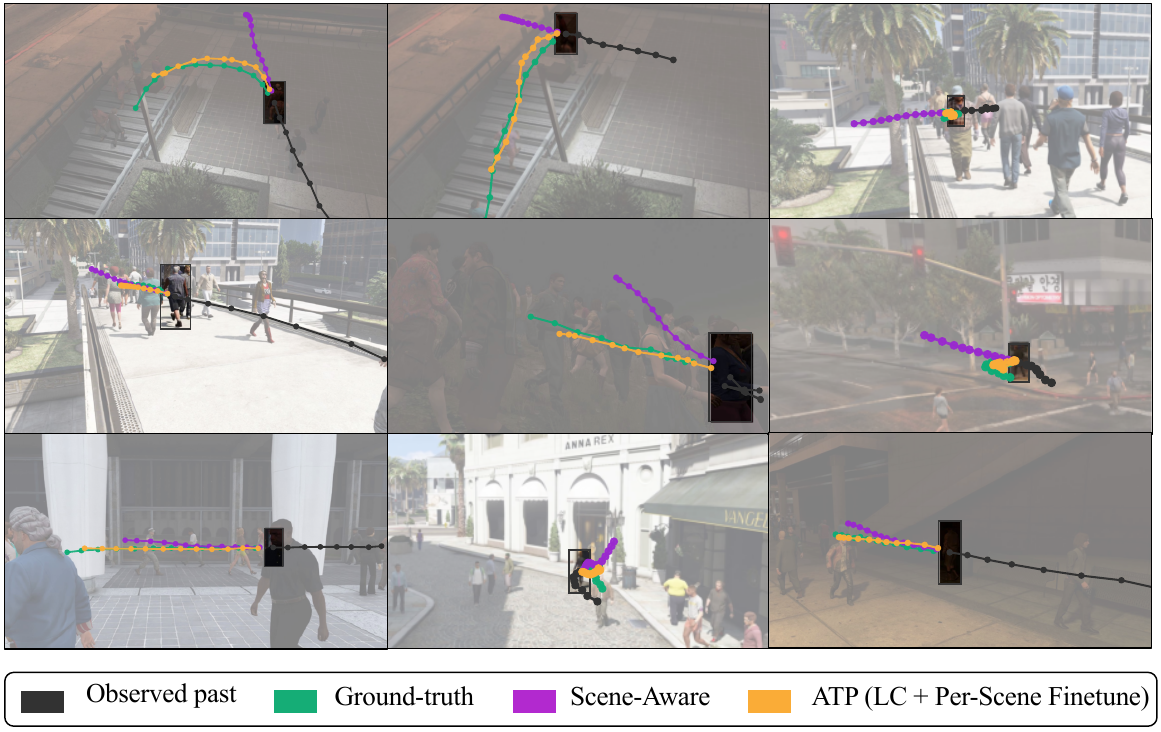}
  \caption{\textbf{Additional MOTSynth qualitative results}. (top left and middle) From several examples of pedestrians in motion, our approach (orange) is able to learn that in this scene, most pedestrians will turn to go down the stairs, while the scene-aware baseline (purple) struggles to understand this scene-specific feature, and instead assumes that the pedestrian will continue walking in the direction of the observed history. Our model is also able to gain understanding that most pedestrians will stay on a walkway, even if they move in a direction orthogonal to it (top right). We also see many more examples of our approach having better awareness of the 3D nature of the ground plane projected into the 2D image (bottom two rows), even when the ground plane is tilted (middle middle).}
  \label{fig:MS-qual-supp}
\end{figure*}

In Fig.~\ref{fig:adaptation_time_FDE}, we see the FDE results for MOTSynth, WildTrack and MOT over time. Similar to results with the ADE metric, we see that on real data (Fig.~\ref{fig:adaptation_time_FDE}a), the adaptive finetuning baseline is slightly better than just latent corridors, but ATP via both latent corridor prompting and per-scene finetuning largely outperforms both of those, and all adaptive methods outperform the non-adaptive baselines. On the MOTSynth data, as with ADE, the ATP via just per-scene finetuning or just latent corridors are comparable (Fig.~\ref{fig:adaptation_time_FDE}b). With FDE, while some scenes have minimal gains, we see even more significant error reduction for some scenes than with ADE, of up to $91.4\%$, and an average FDE improvement of $33.9\%$ from ATP via latent corridors and per-scene finetuning on 25 MOTSynth scenes (Fig.~\ref{fig:adaptation_time_FDE}c).

We showcase additional qualitative results on MOTSynth data in Fig.~\ref{fig:MS-qual-supp}. We see that our approach is able to learn that in a nighttime scene with a large staircase, pedestrians mostly move towards the staircase, regardless of the direction of their observed history (top left and middle). We also see that our approach learns that pedestrians tend to stay on a walkway (top right). Finally, we see several examples of our approach having awareness of the 3D ground plane and predicting future trajectories that lie within the ground plane (bottom two rows).

Quantitative results for each of the EarthCam scenes can be seen in Table~\ref{tab:webcam_quant}. We see that across the four EarthCam scenes, ATP via per-scene finetuning alone works better than using latent corridor adaptation alone (by a narrow margin on both NoLA scenes, and by a significant amount on the Rick's Cafe scene), but a combination of the two is significantly better than any other adaptive or non-adaptive approach. Interestingly, for the Rick's Cafe and Times Square scenes, a constant velocity baseline is better than our non-adaptive learned baselines, but our ATP approach outperforms all baselines.

\begin{table*}[t]
\centering
\begin{tabular}{l|ll|ll|ll|ll}
\textbf{Method} & \multicolumn{2}{c|}{\textbf{Rick's Cafe}} & \multicolumn{2}{c}{\textbf{Times Square}} & \multicolumn{2}{c}{\textbf{NoLA (Day)}} & \multicolumn{2}{c}{\textbf{NoLA (Night)}} \\
 & ADE            & FDE    & ADE           & FDE   & ADE            & FDE    & ADE           & FDE    \\ \hline
Constant velocity     & 9.6 & 16.3 & 15.1  & 26.9 & 36.0   & 70.7 & 48.4& 94.8   \\
Learned Traj (PECNet-Ours) & 20.6 & 29.6  & 42.7 & 60.5  & 35.1 &	64.1& 38.8	&67.8 \\
Scene-aware (YNet-Ours) & 10.4& 16.8  & 17.3 & 30.5  & 30.7	& 59.2 & 37.3	&71.0 \\
ATP (Finetune) & 7.4& 10.6& 14.7 & 25.1& 30.4 & 57.1 & 34.9   & 62.0 \\
ATP (LC) & 10.2   & 16.5   & 16.7  & 29.0  & 30.3 & 58.0 & 36.5    & 67.8  \\
ATP (LC + Per-Scene FT) & \textbf{6.2} & \textbf{8.8} & \textbf{11.7}  & \textbf{19.5} & \textbf{22.6} & \textbf{43.0} & \textbf{29.5} & \textbf{51.9} \\
\end{tabular}
\caption{Results on four EarthCam scenes. Across all of these challenging in-the-wild scenarios, our ATP method using latent corridors and per-scene finetuning outperforms the baselines and other ATP methods.}
\label{tab:webcam_quant}
\end{table*}

\subsection{Results on ETH/UCY}

The main text focused on challenging datasets with non-top-down camera viewpoints, as compared to birds-eye-view datasets popular in earlier works such as SDD and ETH/UCY. Here, we run our approach in this setting by evaluating on ETH/UCY. For each scene in ETH/UCY (ETH, HOTEL, UNIV, ZARA1, ZARA2), we construct an 80/20 train-test split and evaluate as described in main text section 6.2. Results in Table~\ref{tab:eth_ucy} are in pixels and we compute ADE/FDE on a single predicted future. We see a $4.6\%$ improvement in ADE and $2.4\%$ in FDE using LC over per-scene finetuning.

\begin{table}[t]
\centering
\begin{tabular}{lll}
Method                 & ADE           & FDE       \\ \hline
Scene Aware (YNet-Ours)      &  32.3 & 69.1  \\
ATP (Per-Scene Finetune)    & 22.8 & 46.6 \\
ATP (LC + Per-Scene Finetune) & \textbf{21.8}&  \textbf{45.5}\\
\end{tabular}
\caption{Results on ETH/UCY. ATP using latent corridors and per-scene finetuning outperforms ATP using fine-tuning alone, and both approaches successfully adapt over the scene-aware baseline.}
\label{tab:eth_ucy}
\end{table}

\section{Choice of Baselines}
\label{sec:baselines}
Our key scene-aware baseline is YNet, which outperforms or is comparable to other methods that utilize scene priors and trajectory histories such as~\cite{marchetti2020mantra,manh2018scene,xue2018ss,meng2022forecasting}. 
Since we used a simplified version of the YNet architecture for our scene-aware baseline, we have further benchmarked against the original YNet using the codebase training configuration using MOTSynth and the Stanford Drone Dataset~\cite{robicquet2016learning}. We see in Table~\ref{tab:ynet} that the YNet-Ours outperforms the original YNet in the unimodal setting.

\begin{table*}[t]
\centering
\begin{tabular}{l|ll|ll|ll}
 & \multicolumn{2}{c|}{SDD} & \multicolumn{2}{c|}{MOTSynth} \\
Method                 & ADE          & FDE  & ADE          & FDE     \\ \hline
YNet~\cite{mangalam2021goals}    & 24.9        & 49.9 & 54.4          &112.3  \\
YNet-Ours       & \textbf{17.3}          & \textbf{33.6} & \textbf{47.3}          & \textbf{96.5}    \\
\end{tabular}
\caption{The original YNet model compared to our unimodal-only implementation (YNet-Ours). We see that on both SDD and MOTSynth, YNet-Ours outperforms YNet, and is therefore a stronger baseline.}
\label{tab:ynet}
\end{table*}

\end{document}